\documentclass[letterpaper]{article} 
\usepackage[]{aaai24}  
\usepackage{times}  
\usepackage{helvet}  
\usepackage{courier}  
\usepackage[hyphens]{url}  
\usepackage{graphicx} 
\usepackage{natbib}  
\usepackage{caption} 
\usepackage{algorithm}
\usepackage{algorithmic}
\usepackage{tikz}
\usepackage{pgfplots}
\usepackage{amssymb}
\definecolor{cvprblue}{rgb}{0.21,0.49,0.74}

\usepackage{newfloat}
\usepackage{listings}
\DeclareCaptionStyle{ruled}{labelfont=normalfont,labelsep=colon,strut=off} 
\floatstyle{ruled}
\newfloat{listing}{tb}{lst}{}
\floatname{listing}{Listing}
\author {
    Jay Mahajan,
    Samuel Hum,
    Jack Henhapl,
    Diya Yunus,
    Matthew Gadbury,
    Emi Brown,
    Jeff Ginger,
    H. Chad Lane
}

\affiliations {
    \textit{University of Illinois - Urbana Champaign}\\
    \textit{\{ jaym2, hum3, jackah2, dyunus2, gadbury2, ecbrown2, ginger, hclane\}@illinois.edu} 
}

\title{MineObserver 2.0: A Deep Learning \& In-Game Framework for Assessing Natural Language Descriptions of Minecraft Imagery}

\begin{document}

\maketitle

\begin{abstract}
MineObserver 2.0 is an AI framework that uses Computer Vision and Natural Language Processing for assessing the accuracy of learner-generated descriptions of Minecraft images that include some scientifically relevant content. The system automatically assesses the accuracy of participant observations, written in natural language, made during science learning activities that take place in Minecraft. We demonstrate our system working in real-time and describe a teacher support dashboard to showcase observations, both of which advance our previous work. We present the results of a study showing that MineObserver 2.0 improves over its predecessor both in perceived accuracy of the system's generated descriptions as well as in usefulness of the system's feedback. In future work we intend improve system-generated descriptions, give teachers more control and upgrade the system to perform continuous learning to more effectively and rapidly respond to novel observations made by learners.
\end{abstract}

\section{Introduction}
There are few challenges more important to educators and parents than helping children come to a better understanding of the world. Central to this goal is to promote their abilities to articulate what they see and observe about the natural world and to explore those observations through the lens of science \cite{arias2016making}. 

In previous work we presented a prototype AI system called {\it MineObserver} \cite{MineObserver} that sought to assess learner observations in the popular game Minecraft. In this paper, we report on several significant updates to that system (MineObserver 2.0) that improve it along several critical aspects. In the context of video game research, MineObserver 2.0 is a framework that uses machine learning image captioning and a Minecraft plugin that assists users in Minecraft environments designed to promote science learning. The system seeks to both assess learner observations of these worlds and deliver feedback intended to improve their ability to make accurate and productive observations primarily in the context of Astronomy learning.

\section{Background}

\subsection{Minecraft}
Minecraft is an exceptionally popular game. Since its release in 2009, the user base has exploded with over 140M players and 241M logins per month and consistently ranks in the top 5 most popular games for children.\footnote{https://news.xbox.com/en-us/wp-content/uploads/sites/2/2021/04/Minecraft-Franchise-Fact-Sheet\_April-2021.pdf} It spans many platforms and its players have a range of interests, ages and experience. It is referred to as a "sandbox" game because it can be used in several different modes and contexts and often participants come up with their own challenges and meanings when playing alone or with others. The Java Edition of the game has an enormous community following and is very modifiable, which makes it an ideal candidate to create more complex teaching and learning simulations like the one exhibited in this paper.

\subsection{Minecraft as a Learning Environment}
As an extremely modifiable game, Minecraft can be used as a learning environment. Several studies \cite{LANE2017145,MathMinecraft,music,Civic} have used Minecraft to teach a variety of topics including (but not limited to) basic mathematics, civic engagement, and even music. Nearly each study found that Minecraft's flexibility allowed the students to be more creative and have more freedom than a traditional classroom setting. A key component of learning environments is the ability to collaborate with other learners. This is easily achieved since Minecraft can be played with multiple users which can encourages collaboration including collecting difficult achievements, building large structures, and even exploring worlds that are treacherous together. Each of these tasks can be done alone, but would require a great deal of time and effort on one single player. Moreover, modifying the game via a Java plugin, programmers can create new 
tools or achievements that require multiple users in order to achieve a rich collaborative experience.

\subsection{Minecraft for Science Learning}
The research shared in this paper is part of the NSF-funded
project WHIMC (What-If Hypothetical Implementations in Minecraft). WHIMC investigates the use of Minecraft as an educational tool for science learning, with an emphasis on Astronomy content that engages children and promotes interest in STEM \cite{Lane2022Triggering}. WHIMC is implemented as a Minecraft (Java Edition) server consisting of a space station hub and a collection of worlds to visit. On these worlds, learners interactively explore the scientific consequences of alternative versions of Earth via “what if?” questions (e.g., “What if the earth had no moon?”) as well as feasible representations of several known exoplanets. It is hoped that such experiences will act as \textit{triggers} of interest \cite{yi2021identifying}, which are required in order for interest to be sustained over time \cite{renninger2015power}.

\begin{figure}[H]
    \centering
    \includegraphics[width=8.0cm]{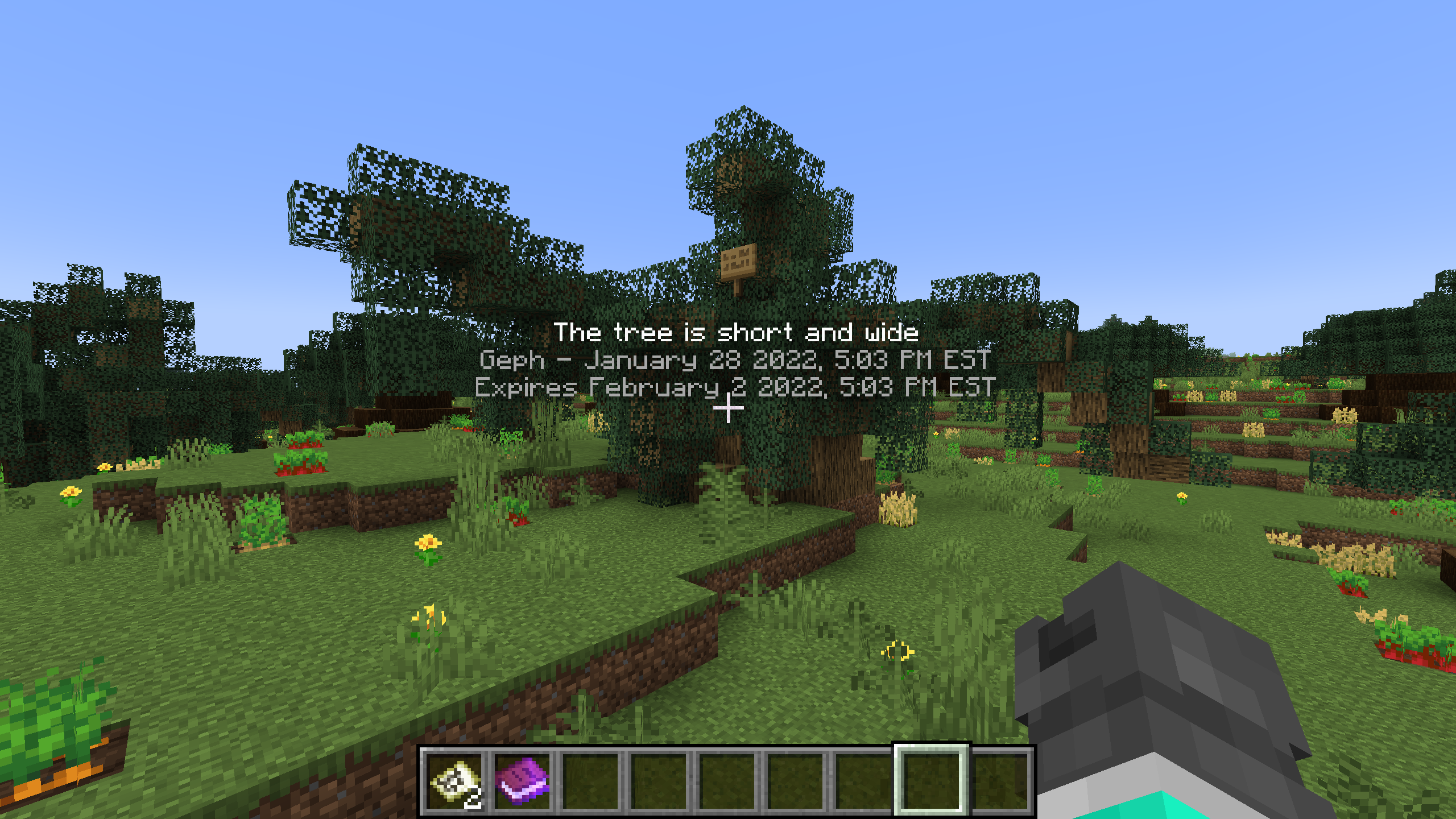}
    \caption{A sample observation of tree variation on a version of Earth with no moon.}
    \label{fig:sample-observation}
\end{figure}

A key component of the project is to analyze how learners interact with the environment and assess their engagement with and understanding of science content. Players are invited to participate in exploration challenges where they can learn about and measure pertinent science characteristics of simulated worlds (such as temperature and radiation). In addition to measurements, learners also make written observations about things they think are noteworthy or interesting in some way, which appear as floating text. For example, without a moon and its gravitational pull, the Earth's rotation would be much faster than it is now. This would cause fierce winds on the surface of Earth. To withstand the force of such wind trees would need to be shorter, wider, and stronger \cite{comins1993if}. An example of an observation for this phenomena is shown in Figure \ref{fig:sample-observation}. We note that the combination of a screenshot and a description forms the basis for our data set described in the next section.

WHIMC provides a framework for making observations like a scientist. In particular, based on our prior work to assess learner observations~\cite{yi2020coding}, we identified five key categories for observations: 

\begin{enumerate}
    \item \textit{Factual}: observations are comprised of nouns without any elaboration.
    \item \textit{Descriptive}: observations related to color, temperature, quantity, and other physical attributes such as weight or size.
    \item \textit{Comparative}: observations comparing one natural phenomenon to another.
    \item \textit{Analogies}: observations comparing natural phenomena with another similar structure or object (an advanced form of comparative). 
    \item \textit{Inferences}: observations where a hypothesis or explanation is proposed (the most advanced form, rare for middle school students to do spontaneously).
\end{enumerate}

Observations are also visible to all players on the server, so they might prompt other learners to take notice. In addition, the WHIMC server captures additional data, including player coordinate and directional facing data to better understand what students were observing at the time.

\subsection{MineObserver}
Previously, MineObserver introduced a AI-method to assess Minecraft imagery from students in a Minecraft world. This method used Image Captioning and cosine similarity to check if a student was correctly observing the intended structure, building, etc. This initial effort was not implemented in-game and instead acted like an outside grader. Moreover, the feedback system lacked depth and did not adequately focus on improving learner observation skills. Our current work improves on this method by directly interacting with the student while improving all aspects of the framework (captioning, feedback, etc.) to give the students a better learning experience and more directly support their emerging skills in making scientific observations.

\section{MineObserver 2.0 AI Framework}
\begin{figure*}[]
    \centering
    \includegraphics[width=10.5cm]{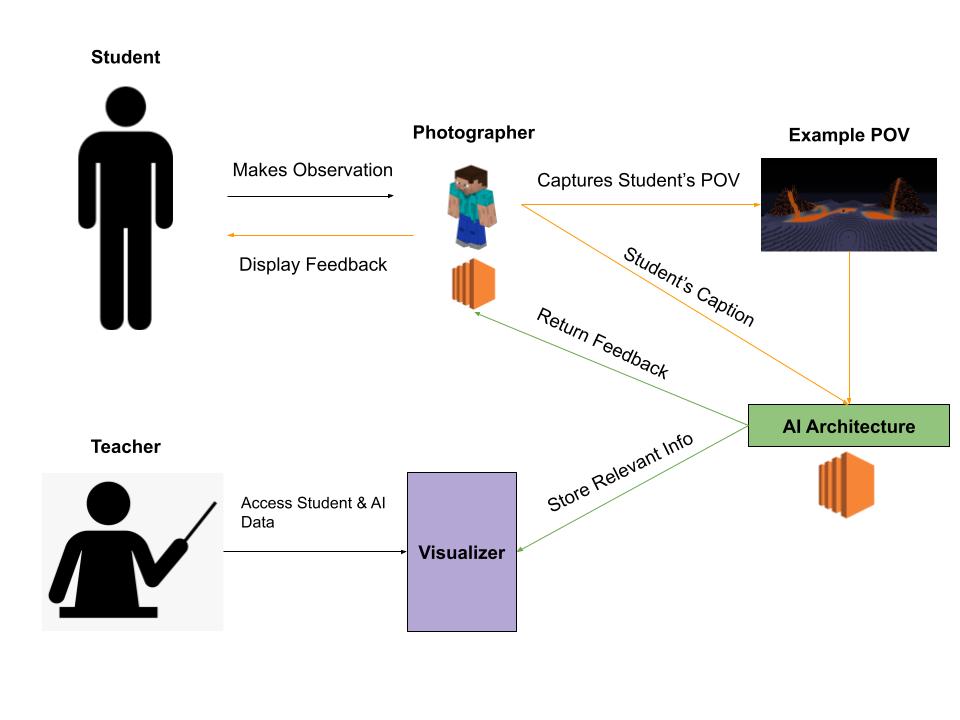}
    \caption{MineObserver 2.0's AI Framework. Note that the Photographer and AI Architecture are hosted on AWS EC2 instances for ease of access.}
    \label{fig:framework}
\end{figure*}

Our AI framework can be broken down into three major parts: the Photographer, the AI Architecture, and the Visualizer. The sequence is as follows. The student makes a visual observation which consists of the student's in-game coordinates and a caption. Next, the Photographer teleports to the student's coordinates and takes a screenshot of the student's POV. The Photographer then sends the student's POV and caption to our AI Architecture. The AI Architecture uses the student's POV and caption to predict how accurate the observation is and gives feedback to the student. The AI Architecture also sends the feedback and a score to the Visualizer. At the end, the student receives feedback via the Photographer and the Visualizer displays the observation made by the student with the AI's feedback. The feedback aims to help an individual student's observation skills while the Visualizer acts an intervention tool for the teacher to further guide a class of students to highlight, focus, and understand which observations are strong, weak, average, etc. Our entire framework can visually depicted in Figure \ref{fig:framework}

\subsection{Photographer}
To enable real-time captioning and feedback in game, we introduce the Photographer. The Photographer is a Minecraft client (a separate account with an independent login) that teleports to a learner's coordinates to capture their point of view via screenshot. This client is tied via unique ID to a spigot-plugin that sends data to our AI Architecture and displays the AI's result to the student in real time, right after they make the observation. To make the Photographer readily available, we host our entire client on an AWS EC2 instance. The student will never physically see the Photographer client in action as it is always in invisible "spectator" mode. This way we avoid interrupting the student's gameplay when making observations.

\subsection{AI Architecture}
\begin{figure}[]
    \centering
    \includegraphics[width=8.0cm]{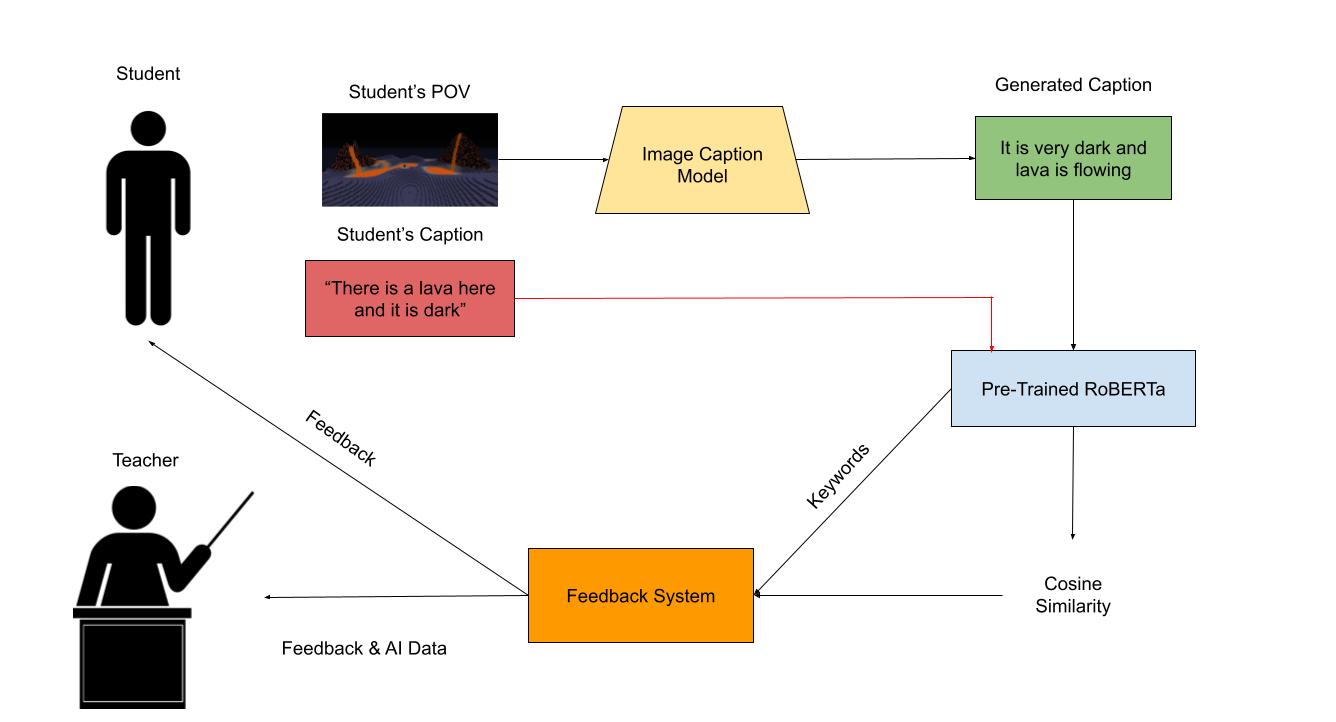}
    \caption{MineObserver 2.0's AI Architecture. }
    \label{fig:ai}
\end{figure}

The AI Architecture consists of three parts: a Image Caption model, a RoBERTa \cite{liu2019roberta} model and the feedback system. From the Photographer, the Image Caption takes the student's POV, we pass the image to a Image Caption model to generate a caption. In this case, the generated caption acts like an expert observer which can be used to compare against the student's observation. Using the student's and generated captions, we pass them through the RoBERTa model. At the end, we use the data from the RoBERTa model to generate feedback and pass the results to the teacher and student. To make the AI Architecture available, we create a REST API and host it on an AWS EC2 instance. We summarize our AI Architecture in Figure \ref{fig:ai}.

\subsection{Image Captioning Model}
Since Minecraft primarily focuses on building and crafting structures, we decided to use an Image Captioning model that can focus on parts of the images. This led us to use an attention based model similar to \cite{showandtell}. In our case, we used a Convolutional Neural Network (CNN) and a Long Short Term Memory (LSTM) \cite{LSTM} with visual attention. While the approach mentioned above uses a pre-trained CNN on ImageNet \cite{imagenet}, we found that this resulted in poor performance thus we train both the CNN and LSTM.

During testing time, instead of using a greedy approach to decode our image, we use beam search with $k = 3$ possible candidates. This allows us to consider the entire sequence rather than individual words when generating our captions.

\subsection{RoBERTa model}
As stated previously, students engaging with the WHIMC platform make observations while exploring different Minecraft maps. These observations are important: they simultaneously reflect how deeply engaged the learner is in the experience and reveal (to an extent) the level of understanding they have for the science concepts. Our agent must assess the content of these observations to guide their pedagogical actions. 

To accomplish this, We utilize the RoBERTa model fine-tuned on the Semantic Textual Similarity benchmark (STSb) and Natural Language Inference (NLI) dataset to encode the image caption generated by the CNN and the student's observation. Facebook's RoBERTa model has been shown to outperform BERT on the General Language Understanding Evaluation (GLUE) benchmark, Stanford Question Answer Dataset (SQuAD), and ReAding Comprehension from Examinations (RACE) dataset~\cite{liu2019roberta}. In addition, fine-tuning the model with STSb and NLI has been shown to improve sentence encodings for common text similarity tasks~\cite{reimers2019sentencebert}. 

For our purposes, we use RoBERTa to perform cosine similarity between the student's caption and the generated caption which produces a score that allows us to understand how closely related the two captions are. We also use this model to perform keyword detection which will be later used in our feedback system.

\subsection{Feedback system}
Using the RoBERTa model, we request $\lambda$ keywords from the generated caption and collect the cosine similarity score. We set a threshold for the cosine similarity $\gamma$ to decide how similar the captions are. If the score is at or above the threshold, then we return the following phase: "Excellent work, you noticed the \{keywords\} here!". And similarly if the score is less than the threshold, then we return the following phase: "Try again! Did you notice the \{keywords\}?". In our case, the keywords acts as primary features in the image to focus on. That way, when the student does not meet the threshold score, our system can guide the student to focus on those parts of the observation and try again. At the end, we send the results back to the student (via Photographer) and the teacher (via Visualizer). 

\subsection{Visualizer}
\begin{figure*}[h]
    \centering
    \includegraphics[width=12cm]{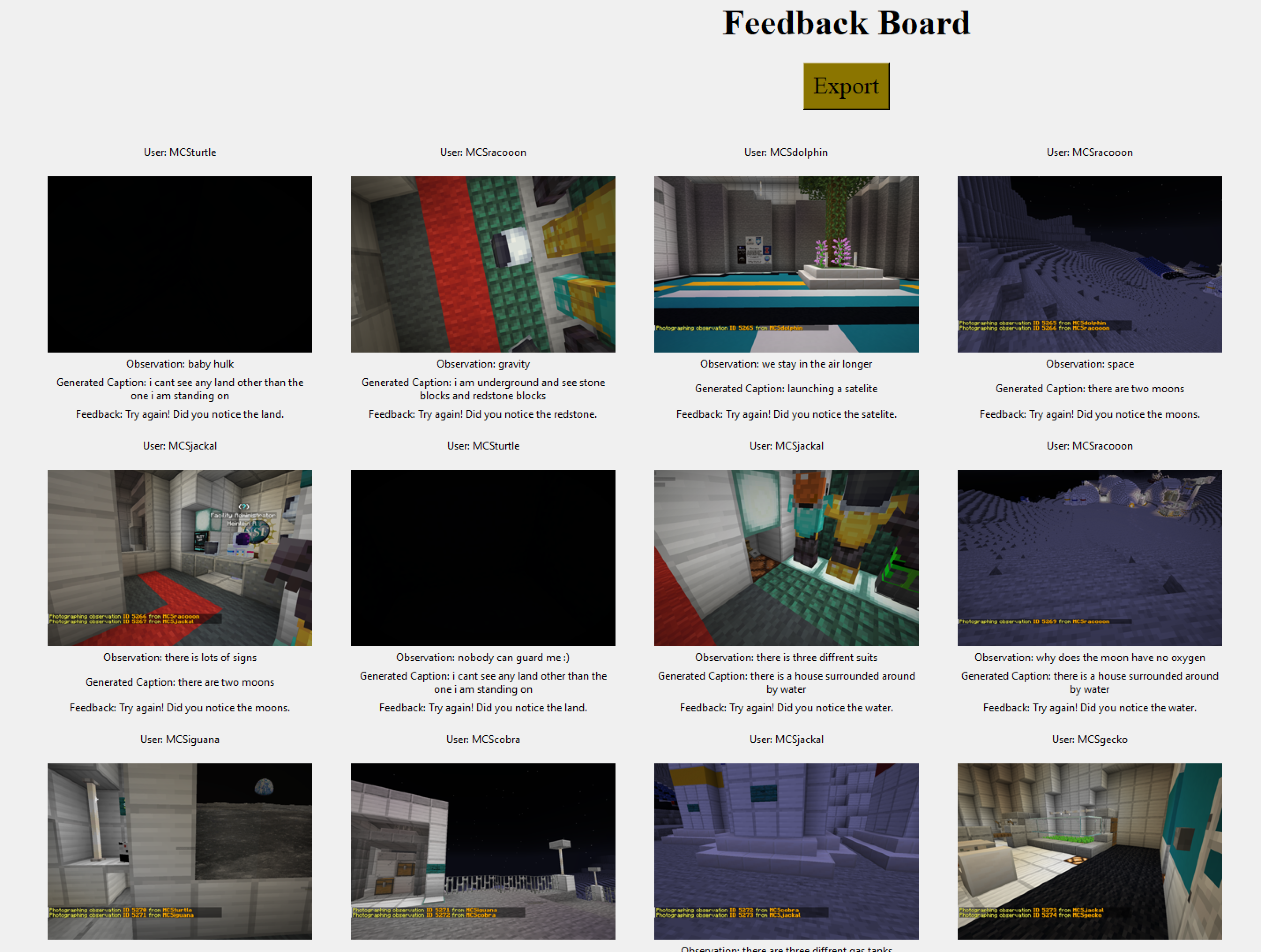}
    \caption{Main UI of the Visualizer. Each of students' observations are listed with their POV, the observation made, the AI's generated caption, and the feedback from the AI's system. At the end, an instructors can export the results to get a folder of images and a CSV of all the Visualizer data}
    \label{fig:vis}
\end{figure*}

The Visualizer is a tool for the instructors to guide students when they make observations. Every time an observation is made, the Visualizer instantly picks it up and displays the student's username, observation, AI-generated observation, and feedback. At the end of the session, the instructor can export all of the data to a folder that contains the images and a CSV of all text data. The goal of the visualizer is to support instructors in initiating discussions with students on observations made and evaluate students' performance and engagement with relevant scientific content. Figure \ref{fig:vis} shows how our Visualizer looks.

\section{Experiments}

\subsection{Dataset}
Since there was no existing dataset of Minecraft images paired with their observations, we had to create our own. We collected hundreds of images of size $1920$ x $1080$ across the WHIMC server and the internet, captioned them using Amazon Mechanical Turk or manually, and labeled the type of observation (e.g Descriptive, Comparative, or Inferential). To target our intended users of this system, we had children from ages 8-14 also caption a set of images and filter out any captions that were either off-topic or extraneous. An important aspect of the WHIMC project is to broaden participation. As such, a substantial amount of our observation training data came from summer camps with participants that are from historically underrepresented groups in STEM.  Eventually, our dataset consisted of over two thousand unique datapoints which was sufficient to capture main points of interest in the WHIMC server.

\subsection{Pre-processing \& Data Augmentation}
Before we load images into our model, we center-cropped them to $1024$ x $1024$ and resized them to be $256$ x $256$ since most of the intended locations of interest lied in the center of the image. When training the model, we augmented our data by adding random horizontal flips and randomly rotated our image by -5 to 5 degrees. 

\subsection{Training}
The only part of our framework that is required to be trained is our Image Caption model. We coded our model using PyTorch \cite{PyTorch}. For our CNN, we decided to use ResNet \cite{ResNet} as our main choice as it is performed against other CNN models such AlexNet \cite{AlexNet} and VGGNet \cite{VGGNet} while being relative light-weight compared to models such as DenseNet \cite{DenseNet}.

\subsection{Implementation}
We trained our entire model via backpropagation by using an Adam optimizer \cite{ADAM}, with a learning rate of 3e-4, and used a cross-entropy loss function for 150 iterations. We used an Nvidia Tesla T4 GPU to train our model quickly. 

For our purposes, we decided to detect $\lambda = 2$ keywords from the generated caption and let the cosine similarity threshold to $\gamma = 0.5$.

\section{Results}

\begin{table*}[h]
\centering

\begin{tabular}{|l|l|l|l|}
\hline
    Significance Test & degrees of freedom & p-value & Significant at $\alpha = 0.05$ \\
    \hline
    \textit{Mean student's rating of AI's generated caption} & 56.806 & \textbf{0.03196} & \checkmark \\
    \hline
    \textit{Mean student's rating of AI's feedback} & 49.521 & \textbf{0.001719} & \checkmark \\
    \hline
\end{tabular}
\caption{Summary of 2-Welch t-test for each experiment. In each case, we compare our model against MineObserver. $n = 61$}
\label{tableWelch}
\end{table*}

\begin{table*}[h]
\centering

\begin{tabular}{|l|l|l|l|l|l|}
\hline
    Model & BLEU-1 & BLEU-2 & BLEU-3 & BLEU-4 & METEOR \\
    \hline
    MineObserver & 0.197 & 0.0294  & \textbf{0.0117} & 0 & 0.12165\\
    \hline
    Ours  & \textbf{0.2} & \textbf{0.033} & 0.0095 & \textbf{0.0055} & \textbf{0.14647} \\
    \hline
\end{tabular}
\caption{Standard Image Captioning Metrics}
\label{tableMetrics}
\end{table*}

\subsection{Summer Camps}

Our tool was used as an integral component of our summer camps, which took place from June until August. A total of 73 participants ages 8 to 14 where ages 8 to 10 made up 40\% and ages 11 to 14 made up 60\% of the student age group. Each camp consisted of roughly 15 hours of instructional time focused on teaching science and astronomy concepts, as well as building in response to habitation requirements. They visited a variety of "what-if" simulations of earth, from "what if earth had two moons" to exoplanet gas giants like Gliese-436b. Along the way students learn about science variables like gravity or atmospheric composition and engage in making scientific observations to later develop and investigate a hypothesis. Many students struggle to know what or how to observe and sometimes get distracted or off-task, our system helps to remedy this. Additional information on our curriculum and approach to teaching and learning with AI and technology can be found at (redacted URL).

For each camp, we conducted a double blind experiment. Each of the camps received either our work or the previous work's framework. Neither did the students nor the instructors knew which work was running at a camp. To ensure a fair comparison, we adjusted the prior work to work at run-time rather after the camp. We conducted this for each of the 6 camps where we explain to the students what AI is via a video and talk and point out that the tool is there to help them. Students first used a simple web interface to each contribute 5 practice observations of still images from our simulation worlds to our future training dataset. They then made a variable amount of observations individually on their own in-game. We summarize the results in table \ref{tableWelch} and in table \ref{tableMetrics}.

\subsection{Accuracy of Generated Captions of Minecraft Images}
We first asked students about the AI's accuracy when generating a caption (observation) to determine if the attention-based image captioning was a useful improvement: 
$$\textit{How accurate was the AI's generated observations?}$$

The students rated the AI's overall accuracy of the generated observations from 1 (not accurate) to 5 (very accurate).

In this case, a 2-sample mean Welch t-test \cite{welch} would be appropriate to compare the students who had our version against the students who had the previous work (MineObserver). Thus, our null and alternative hypothesis can be framed as:


$$H_o: \textit{The mean students' rating of the AI's generated}$$
$\textit{ obervations does not differ between our work and the prior }$
$\textit{work.}$

$$H_a: \textit{The mean students' rating of the AI's generated}$$ $\textit{ obervations for our work is greater than the previous work.}$ $\textit{ }$

If we let $\mu_R$ be the mean students' rating of the AI's generated observation, then mathematically we can expresses this as:
$$H_o: \mu_{Ra} = \mu_{Rb}$$
$$H_a: \mu_{Ra} > \mu_{Rb}$$

Where $a$ is our work and $b$ is the previous work. 
\linebreak

With a degrees of freedom of 56.806, we receive a t-value of 1.8896 which corresponds to a p-value of 0.03196. At $\alpha = 0.05$, this is statistically significant thus we reject the null hypothesis and suggest that the student's mean rating of the AI's generated observations for our work is better than the previous work. This can be seen in table \ref{tableWelch} in our second row.

To further prove that our image captioning model was more accurate, we also computed several pre-existing image captioning metrics. We showcase these in table \ref{tableMetrics}. In almost of all of the metrics, our work beats the previous work.

\subsection{Feedback Accuracy}
Similarly, we asked the students to rate the feedback given by the AI's feedback system. This would ultimately test if students prefer the keyword feedback rather than the generic feedback from the previous work. We posed the following statement to the students:
$$\textit{The feedback from the AI was useful.}$$

The students rated this statement from 1 to 5 where a rating of 1 would correspond to not useful feedback and a rating of 5 would correspond to very useful feedback.

In this case, a 2-sample mean Welch t-test would be appropriate to compare the students who had our version against the students who had previous work. Thus, our null and alternative hypothesis can be framed as:

$$H_o: \textit{The mean students' rating of the AI's feedback does }$$ $\textit{not differ between our work and the prior work.}$

$$H_a: \textit{The mean students' rating of the AI's feedback for }$$ $\textit{our work is greater than the previous work.}$ \linebreak

If we let $\mu_F$ be the mean students' rating of the AI's feedback, then mathematically we can expresses this as:
$$H_o: \mu_{Fa} = \mu_{Fb}$$
$$H_a: \mu_{Fa} > \mu_{Fb}$$

Where $a$ is our work and $b$ is the previous work. 
\linebreak

With a degrees of freedom of 49.521, we receive a t-value of 3.0733 which corresponds to a p-value of 0.001719. At $\alpha = 0.05$, this is statistically significant thus we reject the null hypothesis and suggest that the student's mean rating of the AI's generated observations for our work is better than the previous work. This can be seen table \ref{tableWelch} in our third row.

\section{Other Related Results}

\subsection{In-Game Performance}
\begin{figure}[htp]
    \centering
    \includegraphics[width=8.5cm]{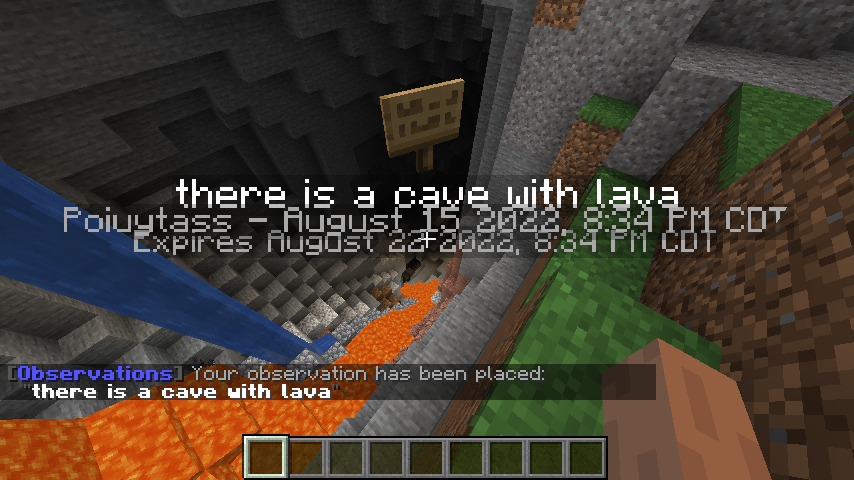}
    \caption{An observation to be graded by MineObserver 2.0's AI }
    \label{fig:sample_obs}
\end{figure}

The experience in game is seamless. A player will make an observation (depicted in figure \ref{fig:sample_obs}) and will get feedback about their caption about 20 seconds later. Once the observation is created the Photographer is put to work. From the student's point of view, they will not see anything; the Photographer is invisible. 

\begin{figure}[htp]
    \centering
    \includegraphics[width=8.5cm]{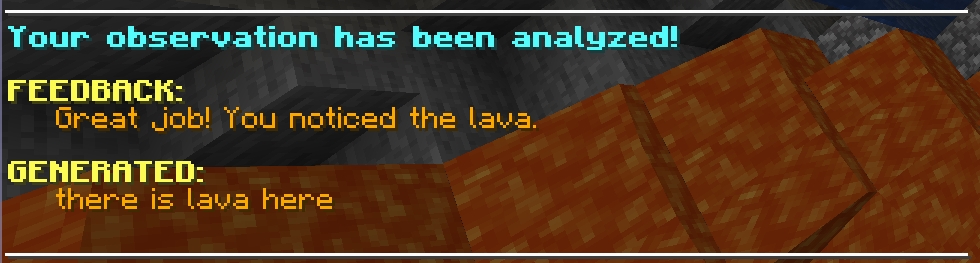}
    \caption{An example of feedback from the AI's feedback system}
    \label{fig:feedback}
\end{figure}

Once the feedback has been generated and returned by the Photographer, the player receives a message in chat with feedback about the player's caption as well as a caption generated by the AI. An example of the response can be seen in figure \ref{fig:feedback}.

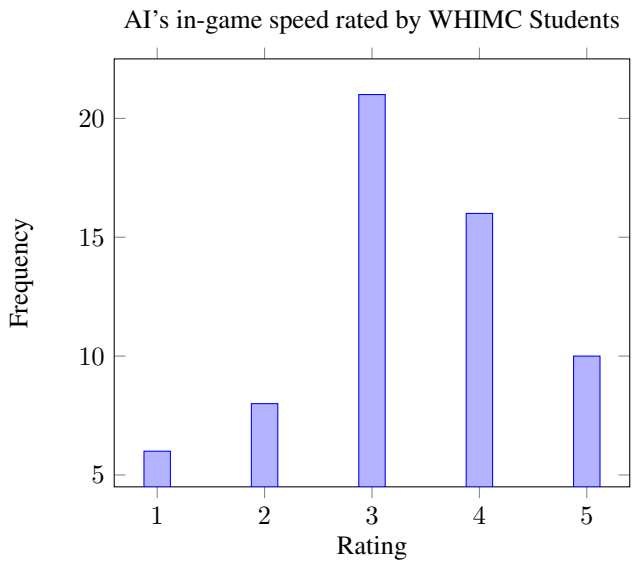
\begin{figure}
    \centering
    \begin{tikzpicture}
    \begin{axis}
    [title = AI's in-game speed rated by WHIMC Students, ybar, ylabel = {Frequency}, xlabel= {Rating}]
        \addplot coordinates {
        (1, 6)
        (2, 8)
        (3, 21)
        (4, 16)
        (5, 10)
        };
        
    \end{axis}
    \end{tikzpicture}
    \caption{Bar Graph of AI's in-game speed rated by WHIMC Summer Camp Students.}
    \label{fig:bar}
\end{figure}

To make sure that the gameplay felt smooth, we asked the students to rate the speed of the AI system from 1 to 5 where a rating of 1 would suggest that the AI was too slow to reply whereas a rating of 5 would suggest that the AI was very quick. Bar chart (figure \ref{fig:bar}) shows the overall result of this question. 

The student gave a median score of 3. This suggest that the in-game speed was adequate but can be improved for a better experience. One suggestion to have more than one Photographer client active to handle the students' activity.

\subsection{Visualizer Results}
Qualitative data was collected via short post-session reflection interviews with five different instructors to assess the effectiveness of using the visualizer dashboard. The instructors relayed that the visualizer was helpful in two major ways. First, it was able to help the them more easily identify which students needed interventions to help stay focused and/or make better observations. This allowed them to focus more on teaching the content of the camps rather than try to visually spot or analyze logs to determine which students were struggling in the middle of session. Second, the instructors used the visualizer to conduct a post-assessment of each day. If a camp had too many students to allow an instructor to pay attention to the dashboard in the moment they could instead leave it on and examine the aggregate visualizer results afterwards. This enabled them to focus on how the overall class performance and reflect on how to address gaps in knowledge or skill during the following camp day. Overall, the instructors unanimously agreed that the visualizer accomplished its primary goal of identifying which students needed intervention in the context of making sufficient observations. That said the interface is still fairly rudimentary and has presents many opportunities for improvement, including (1) cross-platform compatibility via being able to be run as a web app, (2) ability to sort or summarize results for a given time period, (3) the addition of learner profiles to track progress and engagement over time, which could in turn inform the AI, and (4) identify and highlight links that may be common trends or collaboration between participants.

\section{Future Work \& Conclusion}
There are several ideas we wish to add or improve to MineObserver 2.0. This includes styling our image caption model, continuous learning, and stronger feedback.

\subsection{Styling our Image Captioning Model}
Many generative models, such as StyleGAN \cite{styleGAN} allow for generative styles. In our case, we might seek a way to style our image caption model to style based on the type of observation. This would enhance our framework by allowing the student or teacher to choose how they should be graded based on the type of observation.

\subsection{Continuous Learning}
After an observation is made by a student, our framework stores any relevant info to the Visualizer where a teacher can take a look and track student progress. However, since WHIMC is continuously in development, we can use the data from the Visualizer to re-train our image caption model if generated captions are not strong enough. This would allow our model to continually learn as the server grows over time.

\subsection{Stronger Feedback}
Currently, feedback to the learner is limited, typically only directing attention to pertinent visual features nearby and general questions. Future iterations may be able to challenge them with a question to prompt new or additional observations based on their past interests or behaviors, as well as additional engagement with the supporting AI agent or teacher.

\subsection{Additional Applications}
Our tool was developed for the specific context of written observations attached to in-game visuals but may be able to be adapted for alternative use scenarios, including other Minecraft learning simulations or game play. Users could intentionally employ the tool as a kind of note-taking and learner support system while building or exploring. More broadly the ability to take screenshots in a semi-automated fashion based on a Minecraft command interrupt could form the basis of far more complicated pedagogical agents. The plugin and AI essentially mimics and automates the functionality of the camera tool and notebook in Minecraft Education Edition, and could potentially be adapted to create solo-learner self-guided learning experiences. Seeing the great potential we've released our code as open source on Github.

\section{Acknowledgements}
The authors would like to thank the participants and instructors who provided support for the data collection and fellow researchers on the project for help with the coding and statistics. This material is based upon work supported by the National Science Foundation under Grants 1713609 and 1906873.

\bibliography{aaai24}

\end{document}